\documentclass[journal]{IEEEtran}
\usepackage{physics}
\usepackage{graphicx}
\usepackage{float}
\usepackage{amsmath}
\usepackage{subfigure}
\usepackage{amsmath}
\usepackage{longtable}
\usepackage{booktabs}
\usepackage{float}
\usepackage{threeparttable}
\usepackage{xcolor}
\usepackage{diagbox}
\usepackage{cite}
\usepackage{hyperref}
\usepackage{color}
\usepackage{amsfonts}
\usepackage{amsmath}
\usepackage{amssymb}

\usepackage{booktabs}
\usepackage{multirow}

 \renewcommand{\paragraph}[1]{
    \vspace{2mm}
     \noindent\textbf{#1} 
 }
\ifCLASSINFOpdf
\else
\fi

\begin{document}
%
\title{Learning for Vehicle-to-Vehicle Cooperative Perception under Lossy Communication}
%
%
%

\author{Jinlong Li, Runsheng Xu, Xinyu Liu, Jin Ma, Zicheng Chi, Jiaqi Ma, Hongkai Yu$^{*}$ 
\thanks{Jinlong Li, Xinyu Liu, Jin Ma, Zicheng Chi, and Hongkai Yu are with the Department of Electrical Engineering and Computer Science, Cleveland State University, Cleveland, OH 44115, USA.   Runsheng Xu and Jiaqi Ma are with the Department of Civil and Environmental Engineering, University of California, Los Angeles, CA 90024, USA. This work was supported by NSF grants 2215388 and CNS-2127881. This work is part of the OpenCDA Ecosystem~\cite{10045043}.}
\thanks{* Corresponding author: Hongkai Yu (e-mail: h.yu19@csuohio.edu).} 
}


\maketitle 
 \begin{abstract}
Deep learning has been widely used in intelligent vehicle driving  perception systems, such as 3D object detection.
 One promising technique is Cooperative Perception, which leverages Vehicle-to-Vehicle (V2V) communication to share deep learning-based features among vehicles.
However, most cooperative perception algorithms  assume ideal communication and do not consider the impact of Lossy Communication (LC), which is very common in the real world, on feature sharing.
In this paper, we explore the effects of LC on Cooperative Perception and propose a novel approach to mitigate these effects. Our approach includes an LC-aware Repair Network (LCRN) and a V2V Attention Module (V2VAM) with intra-vehicle attention and uncertainty-aware inter-vehicle attention. We demonstrate the effectiveness of our approach on the public OPV2V dataset~(a digital-twin simulated dataset) using point cloud-based 3D object detection.   Our results show that our approach improves detection performance under lossy V2V communication. Specifically, our proposed method achieves a significant improvement in Average Precision compared to the state-of-the-art cooperative perception algorithms, which proves the capability of our approach to effectively mitigate the negative impact of LC and enhance the interaction between the ego vehicle and other vehicles.
The code is available at: \href{https://github.com/jinlong17/V2VLC}{https://github.com/jinlong17/V2VLC}. 

\end{abstract}


\IEEEpeerreviewmaketitle

\section{Introduction}
\IEEEPARstart{H}{ow} to perceive the surrounding  objects precisely in complex real-world scenarios is critical for modern intelligent vehicle research. The accurate perception system (\textit{e.g.}, 3D object detection) is the fundamental base for the next  motion planning and control of the intelligent vehicles, which implies tremendous impacts on the driving safety of intelligent vehicles~\cite{paden2016survey,ma2022verification,cao2022future,9930673,9815132}.  

Because of the  perception limitation of the current individual intelligent vehicle~\cite{chen2022milestones,wang2022gated,9809823}, the cooperative perception of Connected Automated Vehicles (CAV) recently attracted much attention in this research community. Compared to the perception of individual intelligent vehicles, recent studies~\cite{xu2022opv2v,xu2022cobevt,chen2019f} show that cooperative perception of CAV can significantly improve the perception performance by leveraging Vehicle-to-Vehicle (V2V) communication technology for information  sharing. Information sharing through V2V communication is an important technology for CAV cooperative perception, which is utilized to observe a wider range and perceive more occluded objects in the complex traffic environment~\cite{wang2020v2vnet,9857660}. There are three ways for information sharing during V2V communication: (1) sharing raw sensor data as early fusion, (2) sharing intermediate features of the deep learning based detection networks as intermediate fusion, (3) sharing detection results as late fusion. Recent state-of-the-art studies~\cite{xu2022opv2v,xu2022v2x} show that intermediate fusion is the best trade-off between detection accuracy and bandwidth requirement. This paper also focuses on the intermediate fusion during  communication for V2V cooperative perception.     


\begin{figure}[!t]
\centering
\subfigure[Pipeline of V2V Cooperative Perception with lossy communication]{%
  \includegraphics[width=1\columnwidth]{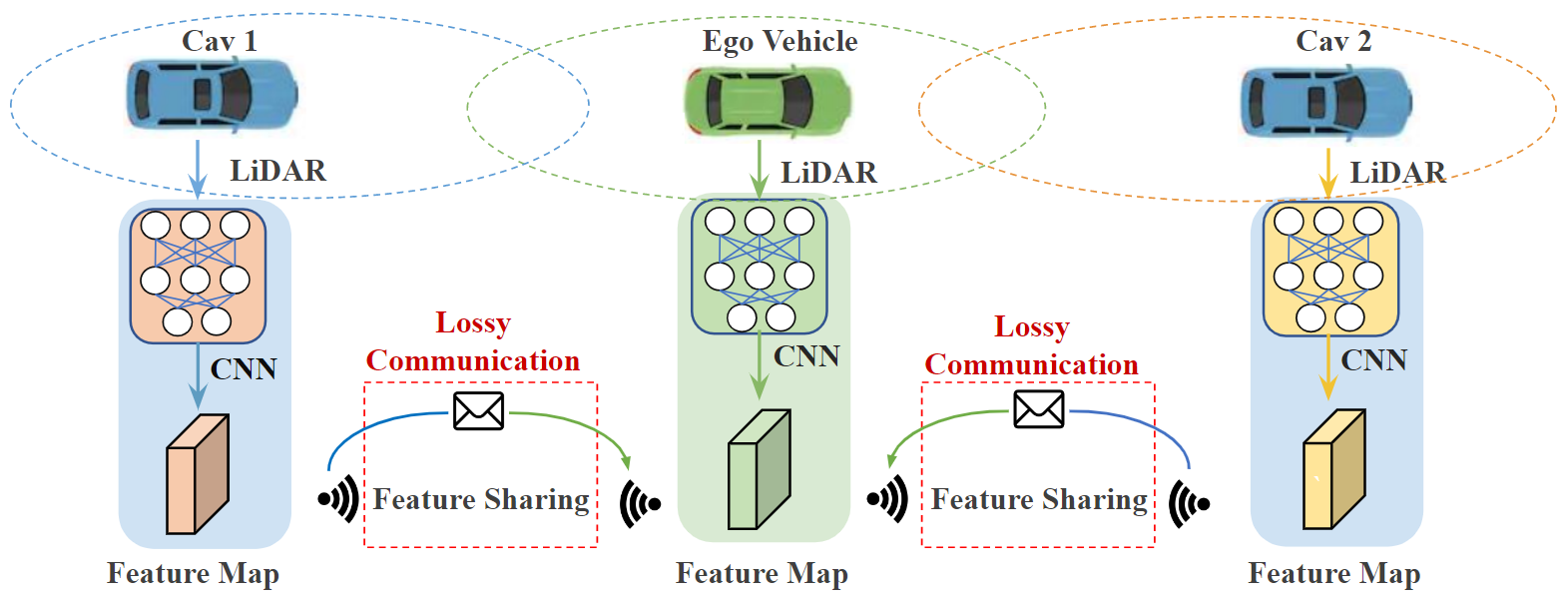}%
}
\hfil
\subfigure[Detection in ideal communication]{%
  \includegraphics[height=0.25\columnwidth,width=0.48\columnwidth]{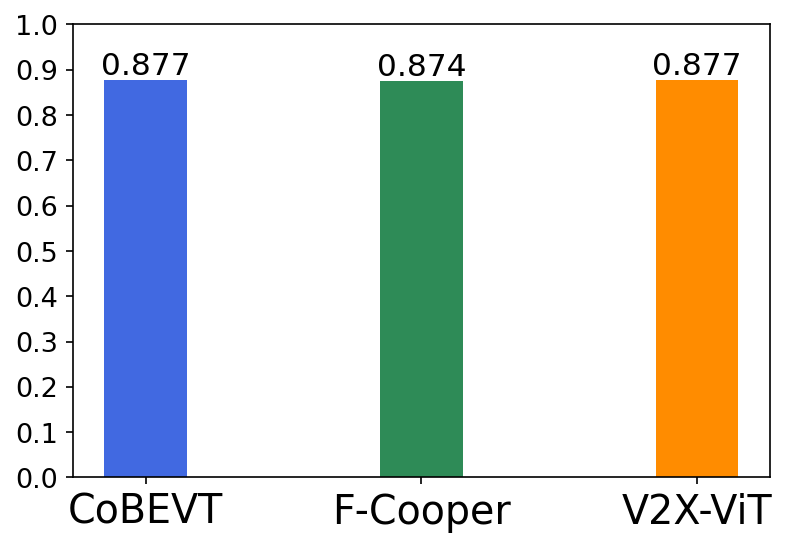}%
}
\hfil
\subfigure[Detection in lossy communication]{%
  \includegraphics[height=0.25\columnwidth,width=0.48\columnwidth]{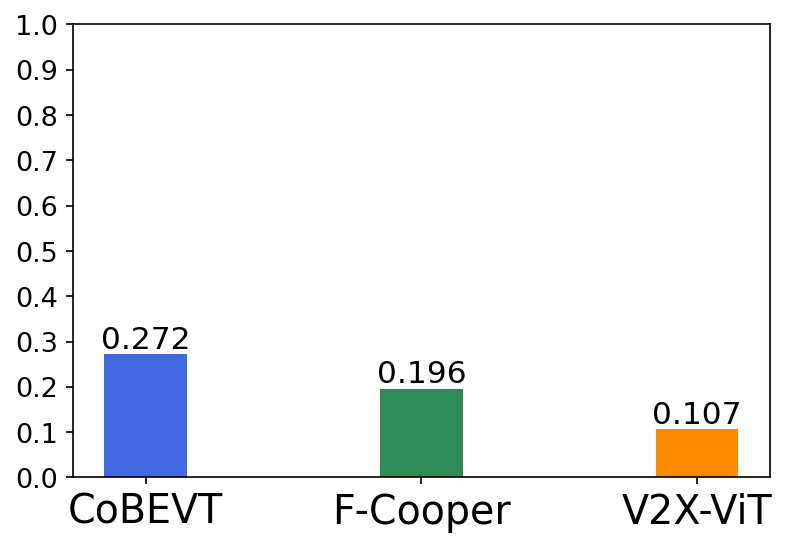}%
}
\caption{Illustration of the V2V cooperative perception pipeline and its detection  performance drop suffering from lossy communication on the public digital-twin CARLA simulator~\cite{dosovitskiy2017carla} based OPV2V dataset~\cite{xu2022opv2v}, where three intermediate fusion methods all trained under ideal communication are displayed:  CoBEVT~\cite{xu2022cobevt}, F-Cooper~\cite{chen2019f}, and V2X-ViT~\cite{xu2022v2x}.}
\label{fig:motivation}
\end{figure}


Many intermediate fusion methods have been recently  proposed for the V2V cooperative perception~\cite{xu2022cobevt,chen2019f,wang2020v2vnet,xu2022opv2v}; however, all of them assume the ideal communication. The only V2V cooperative perception study that considered non-ideal communication focused solely on communication delays~\cite{xu2022v2x}. To date, no existing work has explored the impact of Lossy Communication (LC) on V2V cooperative perception in complex real-world driving environments. 
In urban traffic scenarios, V2V communication is susceptible to a range of factors that can result in lossy communication, such as multi-path effects from obstacles (e.g., buildings and vehicles)\cite{1146527}, Doppler shift introduced by fast-moving vehicles\cite{935159}, interference generated by other communication networks~\cite{10.1145/2988287.2989141}, and dynamic topology caused by routing failures~\cite{10.1145/3427477.3429993},  as well as various weather conditions.
 Incomplete or inaccurate shared intermediate features resulting from lossy communication could compromise the effectiveness and efficiency of V2V cooperative perception, as shown in Figure~\ref{fig:motivation}. Failure to address LC in cooperative perception could lead to degraded perception performance, increased collision risk, and reduced traffic efficiency.

This paper first studies the negative effect of lossy communication in the V2V cooperative perception and then proposes a novel intermediate LC-aware feature fusion method to address the issue.  Specifically, the proposed method includes an LC-aware Repair Network (LCRN) to recover the incomplete shared features by lossy communication and a specially designed V2V Attention Module (V2VAM) to enhance the interaction between the ego vehicle and other vehicles. The V2VAM includes the intra-vehicle attention of ego vehicle and uncertainty-aware inter-vehicle attention. It is challenging to collect the authentic CAV perception data with lossy communication in real-world driving, and considering the advantage of the digital twin in many application~\cite{wang2021digital,wang2020driver,hu2022review,9760104,li2023domain,li2021domain}, this paper evaluates the proposed method in a digital-twin CARLA simulator~\cite{dosovitskiy2017carla} based public cooperative perception dataset OPV2V~\cite{xu2022opv2v}. The contributions of this paper are summarized as follows.

\begin{itemize}
    \item We propose the first research on V2V cooperative perception (point cloud-based 3D object detection) under lossy communication and study the side effect of lossy communication on cooperative perception, specifically the impact on detection performance.
    
    \item This paper proposes a novel intermediate LC-aware feature fusion method to relieve the side effect of lossy communication by a LC-aware Repair Network and enhance the interaction between the ego vehicle and other vehicles by a specially designed V2V Attention Module including intra-vehicle attention of ego vehicle and uncertainty-aware inter-vehicle attention.

    \item We evaluate the proposed method on the public cooperative perception dataset OPV2V, which is based on the digital-twin CARLA simulator~\cite{dosovitskiy2017carla}.
    
\end{itemize}

The rest of this paper is organized as follows. Section~\ref{Sec:Related_Work} briefly reviews the related literature to this work, Section~\ref{Sec:Method} presents the details of the proposed V2V cooperative perception method under lossy communication, Section~\ref{Sec:Experiment} provides the experiments and analysis with two scenarios: Ideal Communication and Lossy Communication, and the final conclusion are given in Section~\ref{Sec:Conclusions}.

\section{Related Work}\label{Sec:Related_Work}

\begin{figure*}[htb] 
    \begin{centering}
        \includegraphics[width=1\textwidth]{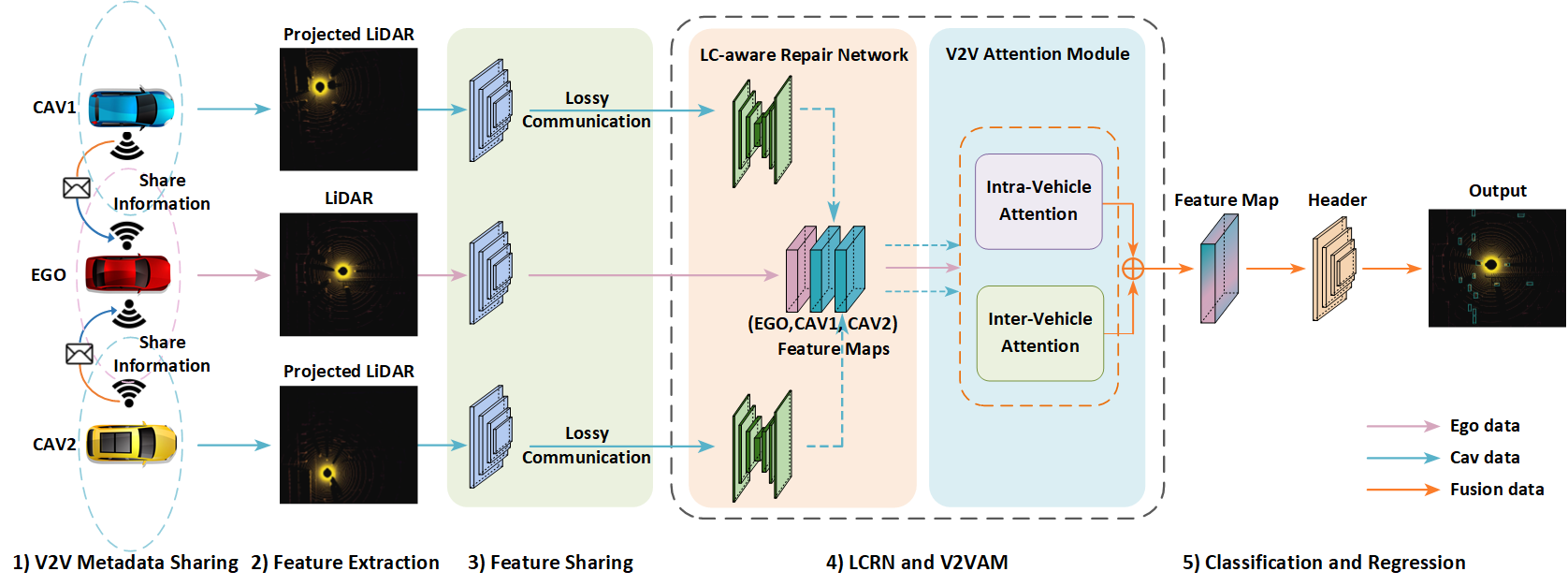}
    \par\end{centering}
    \caption{The architecture of  LC-aware feature fusion framework. The proposed model includes five components: 1) V2V Metadata Sharing, 2) LiDAR Feature Extraction, 3) Feature Sharing, 4) LC-aware Repair Network (LCRN) and V2V Attention Module (V2VAM), 5) Classification and Regression header.}
    \label{fig:architecture}
\end{figure*}

\subsection{3D Perception for Autonomous Driving}

3D object detection is one of  the most critical ways to the success of autonomous driving perception. Based on recently available sensor modality~\cite{9917362}, 3D detection method can be roughly divided into three categories: 
(1) \textbf{Camera-based detection methods} where approaches detect 3D objects using a single or multiple RGB images~\cite{2021Categorical,2021ImVoxelNet,wang2022detr3d}. For example, 
CaDDN~\cite{2021Categorical} utilizes the depth distributions combined with the image features to generate bird’s-eye-view representations for 3D object detection.
ImVoxelNet~\cite{2021ImVoxelNet} constructs a 3D volume in 3D space and samples multi-view features to obtain the voxel representation for 3D object detection. 
DETR3D~\cite{wang2022detr3d} uses queries to index into extracted 2D multi-camera features to directly estimate 3D bounding boxes in 3D spaces.
(2) \textbf{LiDAR-based detection methods} where these methods typically convert LiDAR points into voxels or pillars, leading in voxel-based~\cite{yan2018second,zhou2018voxelnet} or pillar-based object detection methods~\cite{fan2022embracing,wang2020pillar,chen2022density}. 
PointRCNN~\cite{shi2019pointrcnn} proposes a two-stage strategy based on raw point clouds, which learns rough estimation first and then refines it with semantic attributes. Some methods~\cite{yan2018second,zhou2018voxelnet} propose to split the space into voxels and produce features per voxel.
However, 3D voxels are usually expensive to process. To address this issue, PointPillars~\cite{lang2019pointpillars} propose to compress all the voxels along the z-axis into a single pillar, then predict 3D boxes in the bird’s-eye-view space. Moreover, some recent methods~\cite{shi2020pv,yang2019std} combine voxel-based and point-based approaches to detect 3D objects jointly.
(3) \textbf{Camera-LiDAR fusion detection method} where it presents an approach fusing information from both image and LiDAR points, which is a trend in 3D object detection.
How to align the image features with point clouds is challenging in multimodal fusion. To solve this challenge, some methods~\cite{qi2018frustum,vora2020pointpainting,zhou2023bridging} use a two-step framework, where detecting the object in 2D images in the first stage, then using the obtained information to further process point clouds for 3D detection. While other works~\cite{li2022deepfusion,nie2020multimodality} develop end-to-end fusion pipelines and leverage cross-attention mechanisms to perform feature alignment.
Our work  in this paper focuses on the cooperative point cloud based 3D object detection to achieve fast processing and real-time performance~\cite{fan2022embracing,wang2020pillar}; pillar-based approach would be used in our following experiments.

\subsection{Vehicle-to-Vehicle Cooperative Perception}\label{Sec:V2V_related work}

The performance of a 3D perception method highly depends on the accuracy of 3D point clouds. However, LiDAR cameras suffer from refraction, occlusion, and long-range distance, so the single-vehicle system could become unreliable under some challenging situations~\cite{xu2022opv2v}. In recent years, Vehicle-to-Vehicle (V2V) / vehicle-to-infrastructure (V2X) cooperating system has been proposed to overcome the disadvantages of the single-vehicle system by using multiple vehicles. The collaboration among different vehicles enables the 3D perception network to fuse information from different sources.

Some former methods use \textbf{Early fusion} to share raw data among different vehicles. For example, Cooper~\cite{chen2019cooper} fused the point clouds from different connected autonomous vehicles and made predictions based on the aligned data. AUTOCAST~\cite{qiu2021autocast} exchanged sensor readings from different sensors to broaden the perceptive field for a single vehicle. 
Other methods use \textbf{Late fusion} to integrate the 3D detection results from each vehicle. Rawashdeh et al.~\cite{rawashdeh2018collaborative} proposed a machine learning based method that shares the dimension and the location of the center point for each tracked object.
Some other late fusion methods~\cite{arnold2020cooperative,yu2022dair} also adopt point clouds as sensor data from both vehicle and infrastructure. While early fusion requires large bandwidth and data transfer speed, late fusion may generate undesirable results due to the biased individual prediction. In order to find the balance between data load and accuracy, recent methods focus on \textbf{Intermediate fusion} by sharing intermediate representations. F-cooper~\cite{chen2019f} applied voxel features fusion and spatial feature fusion from two cars. V2VNet~\cite{wang2020v2vnet} employed a graph neural network to aggregate features extracted by LiDAR from each vehicle. V2X-ViT~\cite{xu2022v2x} proposed a vision Transformer architecture to fuse features from vehicles and infrastructures.
Cui \textit{et al.}~\cite{cui2022coopernaut} proposed a Point-based Transformer for point cloud processing, which can integrate collaborative perception with control decisions. Tu \textit{et al.}~\cite{tu2021adversarial} proposed an efficient and practical online attack network in a multi-agent deep learning system based on intermediate representations. Luo \textit{et al.}~\cite{luo2022complementarity}  utilized attention modules to fuse the intermediate feature and enhance feature complementarity. Lei \textit{et al.}~\cite{lei2022latency} proposed a latency compensation module to realize intermediate feature-level synchronization. Hu \textit{et al.}~\cite{huwhere2comm} proposed a spatial confidence-aware communication strategy to use less communication to improve performance by focusing on perceptually critical areas.
OPV2V~\cite{xu2022opv2v} utilized a self-attention module to fuse the received intermediate features.
CoBEVT~\cite{xu2022cobevt} proposed local-global sparse attention that captures complex spatial interactions across views and agents to improve the performance of cooperative perception. However, these fusion methods are all with the assumption of ideal communication, which would suffer dramatic performance drop with lossy communication in the real world. To address this issue,  we design a special V2V Attention Module (V2VAM), including intra-vehicle attention of ego vehicle and uncertainty-aware inter-vehicle attention to enhance the V2V interaction.

\subsection{Communication Issue in V2V Perception}
V2V and V2X communication can improve the safety and reliability of autonomous vehicles by exchanging information with surrounding vehicles. However, communication among vehicles may bring new issues to this research field~\cite{zeadally2020tutorial}. 
Due to the nature of the connectivity, lossy communication is inevitable in wireless channels. Some factors like channel errors, network congestion, and delay deadline violation can cause packet losses during the transmission of data in the wireless network~\cite{nasralla2014subjective}.
Low latency and high reliability are two common challenges for V2V communication. For example, in the pre-crash sensing scene, the maximum latency is only 20 ms, and data delivery reliability must be greater than 99$\%$~\cite{zeadally2020tutorial, watta2020vehicle}. Several works have proposed specific resource allocation schemes  to ensure latency and reliability of V2V communication systems by using Lagrange dual decomposition and
binary search ~\cite{mei2018latency}, greedy link selection ~\cite{abbas2018novel}, or federated learning ~\cite{samarakoon2018federated}. Some studies also aim to improve the V2V communication security from different aspects such as authentication, data integrity, and data protection~\cite{hasrouny2017vanet}. 

Lossy Communication (LC) is also a critical issue in V2V communication. According to studies on single-hop broadcasting, the obstacle (vehicles, buildings, etc.) between transmitter and receiver will result in signal power fluctuations, thus causing packet loss~\cite{zeadally2020tutorial, yan2022discrete,sun2021survey, wang2021digital}. The shared data could also be interfered with by other signals or modified by attackers before arriving at its destination, thus leading to lossy data. In this work, we aim to eliminate the lossy communication by proposing an LC-aware repair network and improving the robustness of the V2V perception network.

\section{Methodology}\label{Sec:Method}

In this paper, we focus on the cooperative LIDAR-based 3D object detection task for autonomous driving and consider a realistic scenario where  communication loss exists in collaboration. 
Since we focus on the lossy communication challenge during data transmission in this paper, we assume there are no communication delays or localization errors in the V2V system.
To handle lossy communication challenges in the real world and enhance CAV's cooperative perception capability, inspired by~\cite{xu2022opv2v}, this paper proposes a novel intermediate LC-aware feature fusion framework. The overall architecture of the proposed framework is illustrated in Fig.~\ref{fig:architecture}, which includes five components:
1) V2V metadata sharing, 2) LIDAR feature extraction, 3) Feature sharing, 4) LC-aware repair network and  V2V Attention module, 5) classification and regression headers.

\noindent \subsection{Overview of architecture}

\noindent \textbf{V2V metadata sharing.} We select one of the CAVs as the ego vehicle to construct a spatial graph around it where each node is a CAV within the communication range, and each edge represents a directional V2V communication channel between a pair of nodes. Upon receiving the relative pose and extrinsic of the ego vehicle, all the other CAVs nearby will project their own LiDAR point clouds to the ego vehicle's coordinate frame before feature extraction,
which could be simply formulated as
\begin{equation}
    p_{cav_{projected}}^{t} = T_{cav\rightarrow ego} \cdot p_{cav}^{t},
   \label{eq:tf}
\end{equation}
where $p_{cav}^{t}$  is the CAV pose $[x,y,z,1]^\text{T}$ in $i$-th CAV at the time $t$, and $T_{cav\rightarrow ego}\in \mathbb{R}^{4 \times 4}$ is coordinate transformation matrix from CAV to ego.

\noindent \textbf{LIDAR feature extraction.} The anchor-based PointPillar method~\cite{lang2019pointpillars} is selected as the 3D detection backbone to extract visual features from point clouds. Since it can be deployed in the real world easily than other 3D detection backbones (\textit{e.g.} SECOND~\cite{yan2018second}, PIXOR~\cite{yang2018pixor}, and VoxelNet~\cite{zhou2018voxelnet}) thanks to its low inference latency and optimized memory usage~\cite{xu2022opv2v}. This method converts the raw point clouds to a stacked pillar tensor, then scattered to a 2D pseudo-image and fed to the PointPillar backbone. Finally, the backbone extracts informative visual feature maps. Each CAV has its own LIDAR feature extraction module.

\noindent \textbf{Feature sharing.} In this component, the ego vehicle will receive the neighboring CAV feature maps after each CAV feature extraction, and these received intermediate features will be fed into the remaining detection networks in the ego vehicle. In the real-world scenario (\textit{e.g.} urban building and unpredictable occlusion), the transmission of the feature maps usually suffers inevitable damage that leads to lossy communication. As a result, existing 3D object detectors would suffer a dramatic performance drop with the lossy features collected from surrounding CAVs, as shown in Table~\ref{tab:perfect}. 

\noindent \textbf{LC-aware Repair Network and V2V Attention Module.} The intermediate features aggregated from other surrounding CAVs are fed into the major component of our framework \textit{i.e.}, LC-Aware Repair Network for recovering the intermediate feature map in lossy communication by using tensor-wise filtering, and V2V Attention module for iterative inter-vehicle as well as intra-vehicle feature fusion utilizing attention mechanisms. The proposed LC-aware repair network and  V2V attention module will be revealed with details in Sec.~\ref{Sec:LC-aware} and Sec.~\ref{Sec:V2V}, respectively.

\noindent \textbf{Classification and regression headers.} After receiving the final fused feature maps, two prediction headers are utilized for box regression and classification.

\subsection{LC-aware Repair Network}\label{Sec:LC-aware}

Image denoising is one of the longstanding challenging   tasks in computer vision. The primary sources of noise~\cite{mildenhall2018burst} are shot noise, where a Poisson process with variance equal to the signal level, and read noise, where an approximately Gaussian process is caused by diverse sensor readout effects. To denoise them, some deep learning-based methods~\cite{guo2019toward,niklaus2017video,xia2020basis} use denoising networks that generate a filter for every pixel in the desired output to constrain the output space and thereby prevent the impact of artifacts. Inspired by these architectures, to handle the common V2V communication challenges \textit{i.e.}, lossy communication, we design a customized LC-aware repair network for intermediate feature recovering from other CAVs.

\begin{figure}[htb]  
    \begin{centering}
        \includegraphics[width=0.95\columnwidth]{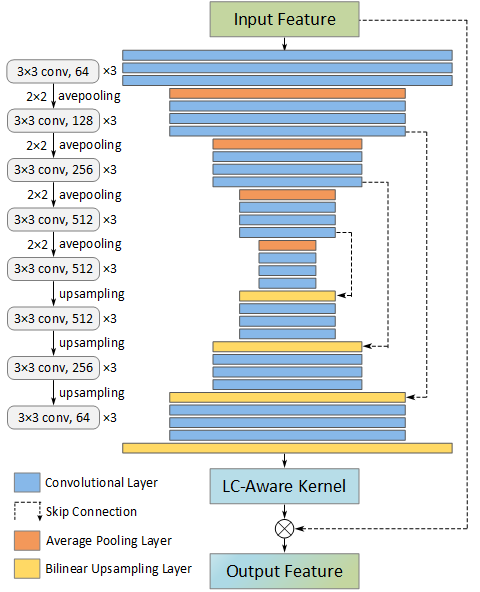}
    \par\end{centering}
    \caption{Illustration of LC-aware Repair Network. The LC-aware Repair architecture for feature recovery is based on the encoder-decoder structure, which outputs per-tensor feature kernels. These kernels then are applied to the input lossy features.}
    \label{fig:loss_repair}
\end{figure}

The framework of the LC-aware repair network is shown in Fig.~\ref{fig:loss_repair},  which is an encoder-decoder architecture with skip connections. This network generates a specific per-tensor filter kernel to jointly align and recover the input damaged feature to produce a recovered version of the output feature.  The input feature for LC-aware repair network is $S\in \mathbb{R}^{c \times h \times w}$, then a tensor-wise kernel $K$ is generated and applied to $S$ to produce the recovered output feature $\hat{S} \in \mathbb{R}^{c \times h \times w}$. the specific tensor-wise filter kernel could be simply formulated as
\begin{equation}
    K = Conv(S),
   \label{eq:kpn_k}
\end{equation}
and the value at each tensor $t$ in our output feature $\hat{S}$ is
\begin{equation}
   \hat{S}^t =  K^t \circledast S^t,
   \label{eq:kpn}
\end{equation}
where $\circledast$ denotes the matrix dot product. $K \in \mathbb{R}^{(k \times k) \times h \times w} $ is a tensor-wise kernel, and each tensor in channel dimension $K^t \in \mathbb{R}^{k \times k}$ is a per-tensor kernel and can be applied to the $k \times k$ neighborhood region of each tensor $t$ in the input feature $S\in \mathbb{R}^{c \times h \times w} $by multiplication. The $Conv(\centerdot)$ denotes the tensor-wise network and is used to perceive the input feature and generate the suitable kernel for each tensor. 

To acquire the repaired output feature $\hat{S}$,  the tensor-wise filtering $\circledast$ of the input damaged feature could largely preserve the feature detailed without corruption. Therefore, a large kernel size $k$ is desired to leverage the rich neighborhood information of each tensor fully. In our experiment, the kernel size $k$ is set to $5$ due to memory limitations.

The LC-aware repair loss function $\mathcal{L}_{LC}(\hat{S},\hat{S}^g)$ is the tensor-wise $L1$ distance between the ground truth original feature $\hat{S}^g$ before suffering lossy communication and the repaired feature $\hat{S}$. The repair loss can be defined as
\begin{equation}
    \mathcal{L}_{LC}(\hat{S},\hat{S}^g) = \lVert \hat{S}^g - \hat{S} \rVert.
   \label{eq:L1}
\end{equation}

\subsection{V2V Attention Module}\label{Sec:V2V}

Self-attention mechanism~\cite{vaswani2017attention} has emerged as a recent advance to capture long-range interaction; The key idea of self-attention is to calculate the response at a position as a weighted sum of the features at all locations, with the interaction between features determined by the features themselves rather than their relative location, as in convolutions. 
In this paper, after receiving the recovered intermediate feature, we aim to leverage the intermediate deep learning features from multiple nearby CAVs to improve perception performance based on V2V communication. We design a customized intra-vehicle and inter-vehicle attention fusion method by considering the lossy communication situation to enhance interaction between ego CAV and other CAVs. Moreover, we adopt a criss-cross attention module in our proposed V2V attention method, which can be leveraged to capture contextual information from full-feature dependencies more efficiently and effectively.

\begin{figure}[htb]  
    \begin{centering}
        \includegraphics[width=1\columnwidth]{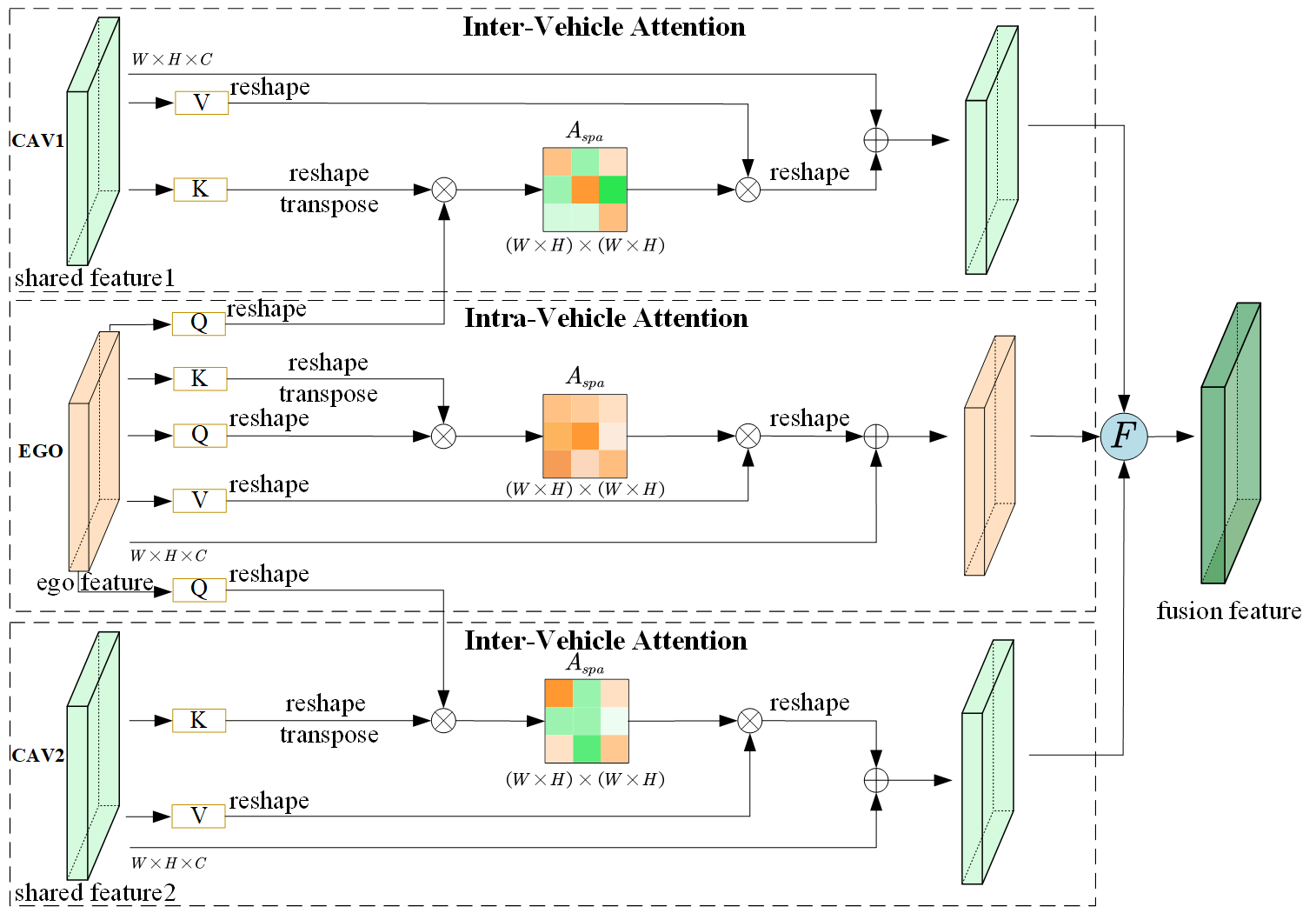}
    \par\end{centering}
    \caption{The architecture of V2V Attention Module includes the intra-vehicle attention of ego vehicle and uncertainty-aware inter-vehicle attention. The final output is a fusion feature with interaction between ego features and other shared features from other CAVs. $\boldsymbol{F}$ denotes a set of max pooling, average pooling, and convolution operation.}
    \label{fig:v2vattention}
\end{figure}

\noindent \textbf{Intra-Vehicle Attention.}\label{intra-vehicle}
For the ego vehicle only, the intra-vehicle attention module can enable  features from any position to perceive globally, thus enjoying full-image contextual information to better capture the representative feature.
Formally, let $\mathbf{H}^{e} \in \mathbb{R}^{C \times H \times W }$ be an input feature map of an ego vehicle, which is perfect data generated by self-vehicle without suffering any lossy communication. In the intra-vehicle attention module, the feature map $\mathbf{H}^{e}$ would be calculated by three $1 \times 1$ convolutional layers to produce three feature vectors $\mathbf{Q}^{e}$, $\mathbf{K}^{e}$, and $\mathbf{V}^{e}$, respectively,  where all of them have the same size, $\{ \mathbf{Q}^{e}, \mathbf{K}^{e}, \mathbf{V}^{e} \} \in \mathbb{R}^{C \times H \times W }$. Following the scaled dot-product attention in~\cite{vaswani2017attention}, we compute the dot products of the $\mathbf{Q}^{e}$ and $\mathbf{K}^{e}$, then divide them using a scaling factor \textit{i.e.} dimension of feature vectors, and apply a softmax function to obtain the weights on the $\mathbf{V}^{e}$. 
The intra-vehicle attention as shown in Fig.~\ref{fig:v2vattention} is defined as follows,
\begin{equation}
    \mathbf{A}^{intra} = {\rm softmax} \left( \frac{\mathbf{Q}^{e} {\mathbf{K}^{e}}^\text{T}}{\sqrt{d^{e}_{k}}} \right) \mathbf{V}^{e}, 
   \label{eq:attention intra}
\end{equation}
where $d^{e}_k$ is the dimension of $\mathbf{K}^{e}$, and the standard $ {\rm softmax}()$ function is used as the   activated function here. $\mathbf{A}^{intra}$ denotes the output feature map of ego vehicle with considering all spatial information of the feature map. 

\noindent \textbf{ Uncertainty-Aware Inter-Vehicle Attention.} 
In V2V cooperative perception, the intermediate feature maps $\mathbf{H}^{s} \in \mathbb{R}^{C \times H \times W}$ aggregated from other CAVs are shared to the ego vehicle. The shared feature maps $\mathbf{H}^{s}$ with lossy communication would be recovered by LC-aware repair network, as introduced in Sec.~\ref{Sec:LC-aware}, but they are still noisy to some extent, while the ego feature maps  $\mathbf{H}^{e}$ are prefect without any lossy transmission. Fusing these uncertain feature maps with a certain ego feature map directly could be risky in the cooperative perception interaction process. To address this issue, we propose an uncertainty-aware inter-vehicle attention fusion method by considering the uncertainty of the recovered feature maps. In this module, the shared feature maps would be calculated by two $1 \times 1$ convolutional layers to produce two feature vectors $\mathbf{K}^{s}$, and $\mathbf{V}^{s}$, respectively,  where all of them have the identical size, $\{ \mathbf{K}^{s}, \mathbf{V}^{s} \} \in \mathbb{R}^{C \times H \times W }$ and the other feature vector $\mathbf{Q}^{e}$ is directly obtained from ego self-vehicle instead of other vehicles, as shown in Fig.~\ref{fig:v2vattention}. Similar to the intra-vehicle attention in Sec.~\ref{intra-vehicle}, the uncertainty-aware inter-vehicle attention can be defined as
\begin{equation}
    \mathbf{A}^{inter} = \sum_{i}^{N} {\rm softmax} \left( \frac{\mathbf{Q}^{e} {\mathbf{K}^{s}_{i}}^\text{T}}{\sqrt{d^{s}_k}} \right) \mathbf{V}^{s}_{i}, 
   \label{eq:attention inter}
\end{equation}
where $d_k$ is the dimension of $\mathbf{K}^{s}_{i}$, and $N$ is the number of the neighboring CAVs.  $\mathbf{A}^{inter}$ denotes the sum of the output feature map  considering the interaction between the ego vehicle and  other vehicles. 

\noindent \textbf{Efficient Implementation.} 
Inspired by~\cite{huang2019ccnet}, we adopt two consecutive criss-cross (CC) attention modules to implement V2V attention in point cloud data rather than scaled dot-product attention. The latter generates huge attention maps to measure the relationships for each point-pair, resulting in a very high complexity of $O(( H \times W)^2)$, where $H \times W$ is the size of input features $\mathbf{H}^{e}$ and $\mathbf{H}^{s}$. The CC attention module~\cite{huang2019ccnet} aggregates contextual information in horizontal and vertical directions, collecting contextual information from all pixels by serially stacking two CC attention modules. Each position has sparse connections to other positions in the feature map, with a total of $(H + W - 1)$ connections per position. This approach greatly reduces the complexity from $O(( H \times W) \times (H \times W ))$ to $O(( H \times W) \times (H + W - 1 ))$ while still effectively capturing the relevant context from all vehicles through V2V communication.

After obtaining the intra-vehicle attention and inter-vehicle attention, all of them would be fed into the max pooling and average pooling layers separately to obtain abundant spatial information, then they are concatenated as the input for the 2D convolutional layer with ReLU activation function.
Therefore, the final fusion feature output $\mathbf{A}^{out}$ in V2V attention module is
\begin{equation}
    \mathbf{A}^{out} = \boldsymbol{F}(\mathbf{A}^{intra} + \mathbf{A}^{inter}),
   \label{eq:attention_sum}
\end{equation}
where $\boldsymbol{F}$ denotes a set of max pooling, average pooling, and convolution layers.
For 3D object detection, we use the smooth L1 loss for bounding box regression and focal loss~\cite{lin2017focal} for classification. The final loss is the combination of detection and LC-aware repair loss $\mathcal{L}_{LC}$ as follows,
\begin{equation}
    \mathcal{L}_{total} = \mu \mathcal{L}_{det} + \lambda \mathcal{L}_{LC},
   \label{eq:total_loss}
\end{equation}
where $\mu$ and $\lambda$ are the balance coefficients within range $[0, 1]$.

\section{Experiment}\label{Sec:Experiment}
\subsection{Dataset}
Due to the difficulties of collecting the real-world CAV perception data for cooperative perception with lossy communication in realistic scenes, we use the digital-twin-based  simulation dataset to validate the proposed method. The experiments are conducted on the public cooperative perception dataset OPV2V~\cite{xu2022opv2v}.
OPV2V is a large-scale open-source simulated dataset for V2V perception, which contains 73 divergent scenes with various numbers of connected vehicles, 11,464 frames, and 232,913 annotated 3D vehicle bounding boxes. These data are collected from 8 digital towns in CARLA~\cite{dosovitskiy2017carla}, and a digital town of Culver City, Los Angeles with the same road topology. Following the default setting of OPV2V~\cite{xu2022opv2v}, we use $3,382$ frames and $1,920$ frames from OPV2V as the training set and validation set, respectively, and $2,170$ frames in CARLA Towns and $594$ frames in Culver City are used as testing set for all methods.

\subsection{Experiments Setup}
\noindent \textbf{Evaluation metrics.} 
We evaluate the performance of our proposed framework by the final 3D vehicle detection accuracy. Following~\cite{xu2022opv2v,xu2022v2x}, we set the evaluation range as $x\in[-140, 140]$ meters,  $y\in[-40, 40]$ meters, where all CAVs are included in this spatial range, whose number is in the range of $[1,5]$ in the experiment. and we measure the accuracy with Average Precisions (AP) at Intersection$-$over$-$Union (IoU) threshold of $0.5$ and $0.7$.

\noindent \textbf{Experiment details.}
In this work, we focus on LiDAR-based vehicle detection and assess models under two scenarios: 1) \textit{Ideal Communication}, where all data transmissions are under perfect communication; 2) \textit{Lossy Communication},  where all intermediate features from other CAVs suffer from the lossy communication except the ego vehicle feature. To simulate the lossy communication, the shared intermediate features are randomly selected by a uniform distributed random probability $p \in [0,1]$, then replaced by a uniform distributed random noise, which is generated by a uniform distribution within the range of original intermediate features. Statistically, the range of original intermediate features is $[0, 29.5]$ in our experiment.

In the training stage, we adopt two schemes to observe the impact of different training data on V2V 3D object detection models. The \textbf{\textit{Scheme I}} uses only ideal communication-based data for training, while the other \textbf{\textit{Scheme II}} uses simulated lossy communication-based data as described above for training. 
The training parameter settings for both schemes are identical, and the only difference between them is the training data, which considers lossy communication in \textbf{\textit{Scheme II}}
All trained models are evaluated on V2V CARLA Towns and Culver City testing sets under both \textit{Ideal Communication} and \textit{Lossy Communication} scenarios. Specifically,  all models use the PointPillar~\cite{lang2019pointpillars} as the backbone with the voxel resolution of $0.4$ m for both height and width. We adopt Adam optimizer~\cite{loshchilov2017decoupled} with an initial learning rate of $10^{-3}$ and steadily decay it every $10$ epochs using a factor of $0.1$.
The coefficient of detection loss $\mathcal{L}_{det}$ is set to $1.0$, and that of LC-aware repair loss $\mathcal{L}_{LC}$ is set to $0.1$. We follow the same hyperparameters in V2X-ViT~\cite{xu2022v2x}, and all models are trained on two RTX 3090 GPUs.

\noindent\textbf{Compared methods.} We consider \textit{No fusion} as the baseline, which only uses the ego vehicle's LiDAR point clouds. In addition, we evaluate five state-of-the-art methods in this paper, which use \textit{Intermediate Fusion} as the main fusion strategy: CoBEVT~\cite{xu2022cobevt}, F-Cooper~\cite{chen2019f}, V2VNet~\cite{wang2020v2vnet}, OPV2V~\cite{xu2022opv2v}, and V2X-ViT~\cite{xu2022v2x}(see Sec.\ref{Sec:V2V_related work} for detailed descriptions). To demonstrate the significant effect of Lossy Communication, we first train these methods under two scenarios: \textit{Ideal Communication} and \textit{Lossy Communication}. We then test these methods under the same two scenarios to assess their performance. To show the effectiveness of two critical components in our framework, namely LCRN, and V2VAM, we design a simple feature averaging fusion method with a $1 \times 1$ convolutional layer called AveFuse. This method averages all intermediate features from ego-vehicle and other vehicles, and then the averaged feature is passed through a $1 \times 1$ convolutional layer.

\begin{table}[]
\caption{3D Object detection performance comparison on two testing sets of OPV2V  based the training of \textit{Scheme I}. We show Average Precision (AP) at IoU=0.5, 0.7. Note that V2VAM is only our proposed V2V  attention module while V2VAM+LCRN is our full Proposed method.}
\label{tab:perfect}
\begin{tabular}{@{}cccccc@{}}
\toprule
\multirow{2}{*}{Method} &
  \multirow{2}{*}{\begin{tabular}[c]{@{}c@{}}Com.\\  Type\end{tabular}} &
  \multicolumn{2}{c}{V2V CARLA Towns} &
  \multicolumn{2}{c}{V2V  Culver City} \\
 &
   &
  AP@ 0.5 &
  AP@ 0.7 &
  AP@ 0.5 &
  AP@ 0.7 \\ \midrule
NO Fusion &
  \multicolumn{1}{c|}{\begin{tabular}[c]{@{}c@{}}Ideal\\ Lossy\end{tabular}} &
  \begin{tabular}[c]{@{}c@{}}0.679\\ 0.679\end{tabular} &
  \multicolumn{1}{c|}{\begin{tabular}[c]{@{}c@{}}0.602\\  0.602\end{tabular}} &
  \begin{tabular}[c]{@{}c@{}}0.557\\ 0.557\end{tabular} &
  \begin{tabular}[c]{@{}c@{}}0.471\\  0.471\end{tabular} \\ \midrule
F-Cooper~\cite{chen2019f} &
  \multicolumn{1}{c|}{\begin{tabular}[c]{@{}c@{}}Ideal\\ Lossy\end{tabular}} &
  \begin{tabular}[c]{@{}c@{}}0.844\\ 0.036\end{tabular} &
  \multicolumn{1}{c|}{\begin{tabular}[c]{@{}c@{}}0.743\\ 0.029\end{tabular}} &
  \begin{tabular}[c]{@{}c@{}}0.874\\ 0.196\end{tabular} &
  \begin{tabular}[c]{@{}c@{}}0.715\\ 0.146\end{tabular} \\ \midrule
V2VNet~\cite{wang2020v2vnet} &
  \multicolumn{1}{c|}{\begin{tabular}[c]{@{}c@{}}Ideal\\ Lossy\end{tabular}} &
  \begin{tabular}[c]{@{}c@{}}0.874\\ 0.024\end{tabular} &
  \multicolumn{1}{c|}{\begin{tabular}[c]{@{}c@{}}0.712\\ 0.014\end{tabular}} &
  \begin{tabular}[c]{@{}c@{}}0.855\\ 0.102\end{tabular} &
  \begin{tabular}[c]{@{}c@{}}0.630\\ 0.061\end{tabular} \\ \midrule
OPV2V~\cite{xu2022opv2v} &
  \multicolumn{1}{c|}{\begin{tabular}[c]{@{}c@{}}Ideal\\ Lossy\end{tabular}} &
  \begin{tabular}[c]{@{}c@{}}0.871\\ 0.015\end{tabular} &
  \multicolumn{1}{c|}{\begin{tabular}[c]{@{}c@{}}0.793\\ 0.011\end{tabular}} &
  \begin{tabular}[c]{@{}c@{}}0.868\\ 0.010\end{tabular} &
  \begin{tabular}[c]{@{}c@{}}0.745\\ 0.006\end{tabular} \\ \midrule
CoBEVT~\cite{xu2022cobevt} &
  \multicolumn{1}{c|}{\begin{tabular}[c]{@{}c@{}}Ideal\\ Lossy\end{tabular}} &
  \begin{tabular}[c]{@{}c@{}}0.914\\ 0.089\end{tabular} &
  \multicolumn{1}{c|}{\begin{tabular}[c]{@{}c@{}}0.836\\ 0.069\end{tabular}} &
  \begin{tabular}[c]{@{}c@{}}0.877\\ 0.272\end{tabular} &
  \begin{tabular}[c]{@{}c@{}} 0.748\\ 0.202\end{tabular} \\ \midrule
V2X-ViT~\cite{xu2022v2x} &
  \multicolumn{1}{c|}{\begin{tabular}[c]{@{}c@{}}Ideal\\ Lossy\end{tabular}} &
  \begin{tabular}[c]{@{}c@{}}0.840\\ 0.083\end{tabular} &
  \multicolumn{1}{c|}{\begin{tabular}[c]{@{}c@{}}0.726\\ 0.054\end{tabular}} &
  \begin{tabular}[c]{@{}c@{}}0.877\\ 0.107\end{tabular} &
  \begin{tabular}[c]{@{}c@{}}0.720\\ 0.067\end{tabular} \\ \midrule
\begin{tabular}[c]{@{}c@{}} V2VAM \end{tabular} &
  \multicolumn{1}{c|}{\begin{tabular}[c]{@{}c@{}}Ideal\\ Lossy\end{tabular}} &
  \begin{tabular}[c]{@{}c@{}} \textbf{0.926}\\ 0.085 \end{tabular} &
  \multicolumn{1}{c|}{\begin{tabular}[c]{@{}c@{}} \textbf{0.861}\\ 0.075\end{tabular}} &
  \begin{tabular}[c]{@{}c@{}} \textbf{0.885}\\ 0.095\end{tabular} &
  \begin{tabular}[c]{@{}c@{}} \textbf{0.785}\\ 0.070\end{tabular} \\ \bottomrule
\end{tabular}
\end{table}


\begin{table}[htb]
\centering
\caption{3D detection performance comparison on two testing sets of OPV2V based on the training of \textit{Scheme II}. }
\label{tab:LC}
\begin{tabular}{@{}cccccc@{}}
\toprule
\multirow{2}{*}{Method} &
  \multirow{2}{*}{\begin{tabular}[c]{@{}c@{}}Com.\\  Type\end{tabular}} &
  \multicolumn{2}{c}{V2V CARLA Towns} &
  \multicolumn{2}{c}{V2V Culver City} \\
 &
   &
  AP@0.5 &
  AP@0.7 &
  AP@0.5 &
  AP@0.7 \\ \midrule
NO Fusion &
  \multicolumn{1}{c|}{\begin{tabular}[c]{@{}c@{}}Ideal\\ Lossy\end{tabular}} &
  \begin{tabular}[c]{@{}c@{}}0.679\\ 0.679\end{tabular} &
  \multicolumn{1}{c|}{\begin{tabular}[c]{@{}c@{}}0.602\\  0.602\end{tabular}} &
  \begin{tabular}[c]{@{}c@{}}0.557\\ 0.557\end{tabular} &
  \begin{tabular}[c]{@{}c@{}}0.471\\  0.471\end{tabular} \\ \midrule
F-Cooper~\cite{chen2019f} &
  \multicolumn{1}{c|}{\begin{tabular}[c]{@{}c@{}}Ideal\\ Lossy\end{tabular}} &
  \begin{tabular}[c]{@{}c@{}}0.677\\ 0.677\end{tabular} &
  \multicolumn{1}{c|}{\begin{tabular}[c]{@{}c@{}}0.494\\ 0.492\end{tabular}} &
  \begin{tabular}[c]{@{}c@{}}0.758\\ 0.656\end{tabular} &
  \begin{tabular}[c]{@{}c@{}}0.523\\ 0.440\end{tabular} \\ \midrule
V2VNet~\cite{wang2020v2vnet} &
  \multicolumn{1}{c|}{\begin{tabular}[c]{@{}c@{}}Ideal\\ Lossy\end{tabular}} &
  \begin{tabular}[c]{@{}c@{}}0.713\\ 0.714\end{tabular} &
  \multicolumn{1}{c|}{\begin{tabular}[c]{@{}c@{}}0.465\\ 0.465\end{tabular}} &
  \begin{tabular}[c]{@{}c@{}}0.702\\ 0.702\end{tabular} &
  \begin{tabular}[c]{@{}c@{}}0.408\\ 0.409\end{tabular} \\ \midrule
OPV2V~\cite{xu2022opv2v} &
  \multicolumn{1}{c|}{\begin{tabular}[c]{@{}c@{}}Ideal\\ Lossy\end{tabular}} &
  \begin{tabular}[c]{@{}c@{}}0.804\\ 0.739\end{tabular} &
  \multicolumn{1}{c|}{\begin{tabular}[c]{@{}c@{}}0.645\\ 0.603\end{tabular}} &
  \begin{tabular}[c]{@{}c@{}}0.742\\ 0.718\end{tabular} &
  \begin{tabular}[c]{@{}c@{}}0.576\\ 0.561\end{tabular} \\ \midrule
CoBEVT~\cite{xu2022cobevt} &
  \multicolumn{1}{c|}{\begin{tabular}[c]{@{}c@{}}Ideal\\ Lossy\end{tabular}} &
  \begin{tabular}[c]{@{}c@{}}0.871\\ 0.768\end{tabular} &
  \multicolumn{1}{c|}{\begin{tabular}[c]{@{}c@{}}0.740\\ 0.582\end{tabular}} &
  \begin{tabular}[c]{@{}c@{}}0.866\\ 0.795\end{tabular} &
  \begin{tabular}[c]{@{}c@{}}0.688\\ 0.586\end{tabular} \\ \midrule
V2X-ViT~\cite{xu2022v2x} &
  \multicolumn{1}{c|}{\begin{tabular}[c]{@{}c@{}}Ideal\\ Lossy\end{tabular}} &
  \begin{tabular}[c]{@{}c@{}}0.793\\ 0.770\end{tabular} &
  \multicolumn{1}{c|}{\begin{tabular}[c]{@{}c@{}}0.619\\ 0.599\end{tabular}} &
  \begin{tabular}[c]{@{}c@{}}0.731\\ 0.717\end{tabular} &
  \begin{tabular}[c]{@{}c@{}}0.520\\ 0.511\end{tabular} \\ \midrule
V2VAM+LCRN &
  \multicolumn{1}{c|}{\begin{tabular}[c]{@{}c@{}}Ideal\\ Lossy\end{tabular}} &
  \begin{tabular}[c]{@{}c@{}} \textcolor{blue}{\textbf{0.887}}\\ \textcolor{red}{\textbf{0.841}} \end{tabular} &
  \multicolumn{1}{c|}{\begin{tabular}[c]{@{}c@{}} \textcolor{blue}{\textbf{0.783}}\\ \textcolor{red}{\textbf{0.705}} \end{tabular}} &
  \begin{tabular}[c]{@{}c@{}} \textcolor{blue}{\textbf{0.871}}\\ \textcolor{red}{\textbf{0.846}} \end{tabular} &
  \begin{tabular}[c]{@{}c@{}} \textcolor{blue}{\textbf{0.709}}\\ \textcolor{red}{\textbf{0.663}} \end{tabular} \\ \bottomrule
\end{tabular}
\end{table}


\begin{table}[htb]
\centering
\caption{Ablation study for 3D object detection on two testing sets of OPV2V based on training of \textit{Scheme II}. Note that V2VAM+LCRN is our Proposed method.}
\label{tab:ablation_result}
\begin{tabular}{@{}ccccc@{}}
\toprule
\multirow{2}{*}{Method} & \multicolumn{2}{c}{V2V CARLA Towns} & \multicolumn{2}{c}{V2V Culver City} \\
                                        & AP@0.5 & AP@0.7                     & AP@0.5 & AP@0.7 \\ \midrule
\multicolumn{1}{c|}{NO Fusion}          & 0.679  & \multicolumn{1}{c|}{0.602} & 0.557  & 0.471  \\ \midrule
\multicolumn{1}{c|}{AveFuse (Baseline)} & 0.632    & \multicolumn{1}{c|}{0.325}   & 0.697    & 0.374    \\ \midrule
\multicolumn{1}{c|}{V2VAM \textit{w/o} Intra}              & 0.613   & \multicolumn{1}{c|}{0.490}   & 0.637    & 0.458    \\ \midrule
\multicolumn{1}{c|}{{V2VAM \textit{w/o} Inter}}            & 0.641   & \multicolumn{1}{c|}{0.494}   & 0.681    & 0.504   \\ \midrule
\multicolumn{1}{c|}{V2VAM}              & 0.709    & \multicolumn{1}{c|}{0.583}   & 0.761    & 0.541    \\ \midrule
\multicolumn{1}{c|}{AveFuse+LCRN}      & 0.698    & \multicolumn{1}{c|}{0.472}   & 0.714    & 0.558    \\ \midrule
\multicolumn{1}{c|}{V2VAM+LCRN}        & \textbf{0.841}    & \multicolumn{1}{c|}{\textbf{0.705}}   & \textbf{0.846}   & \textbf{0.663}    \\ \bottomrule
\end{tabular}
\end{table}

\begin{figure*}[!t]
\centering
\subfigure[F-Cooper~\cite{chen2019f}]{%
  \includegraphics[width=0.65\columnwidth]{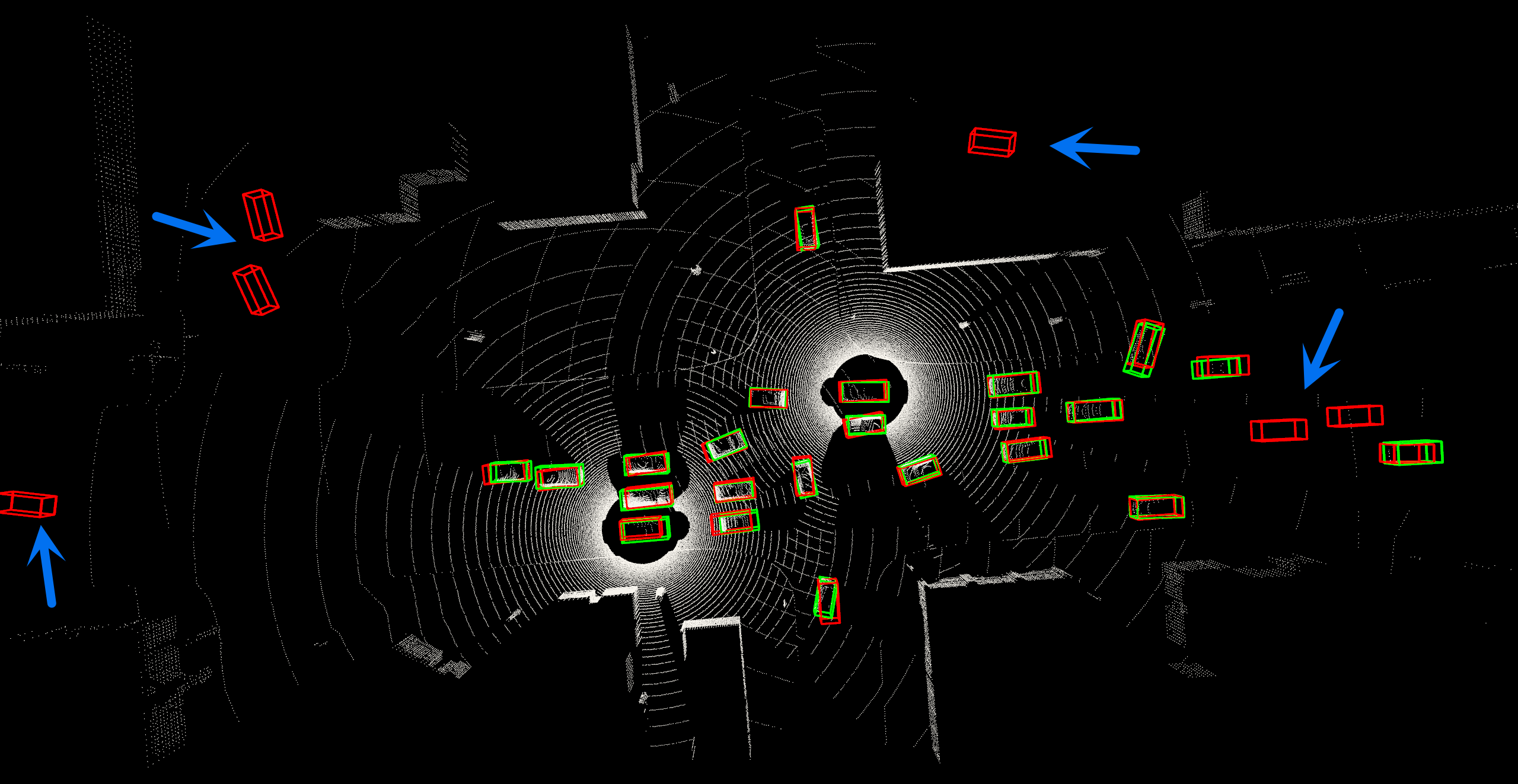}%
}
\subfigure[V2VNet~\cite{wang2020v2vnet}]{%
  \includegraphics[width=0.65\columnwidth]{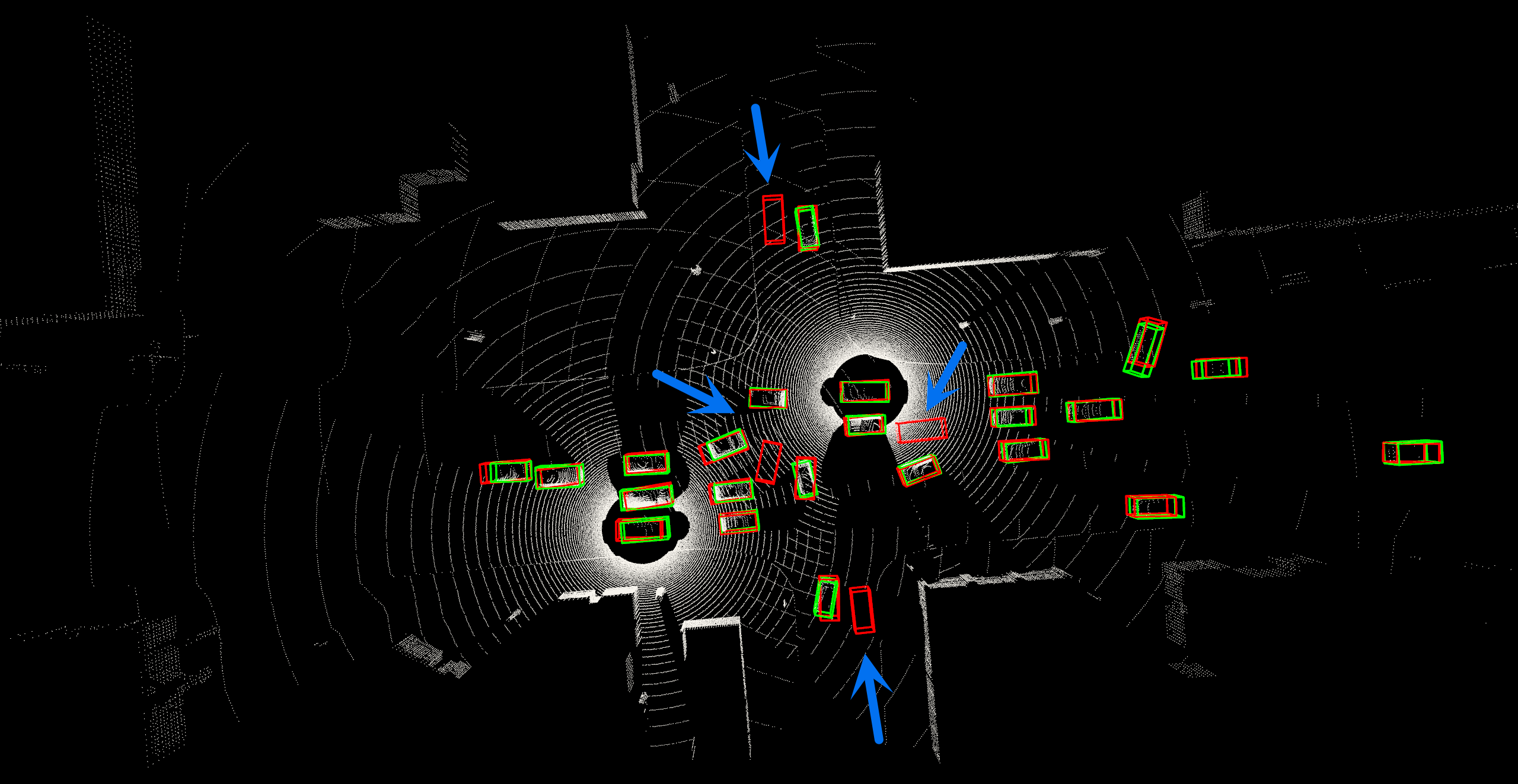}%
}
\subfigure[OPV2V~\cite{xu2022opv2v}]{%
  \includegraphics[width=0.65\columnwidth]{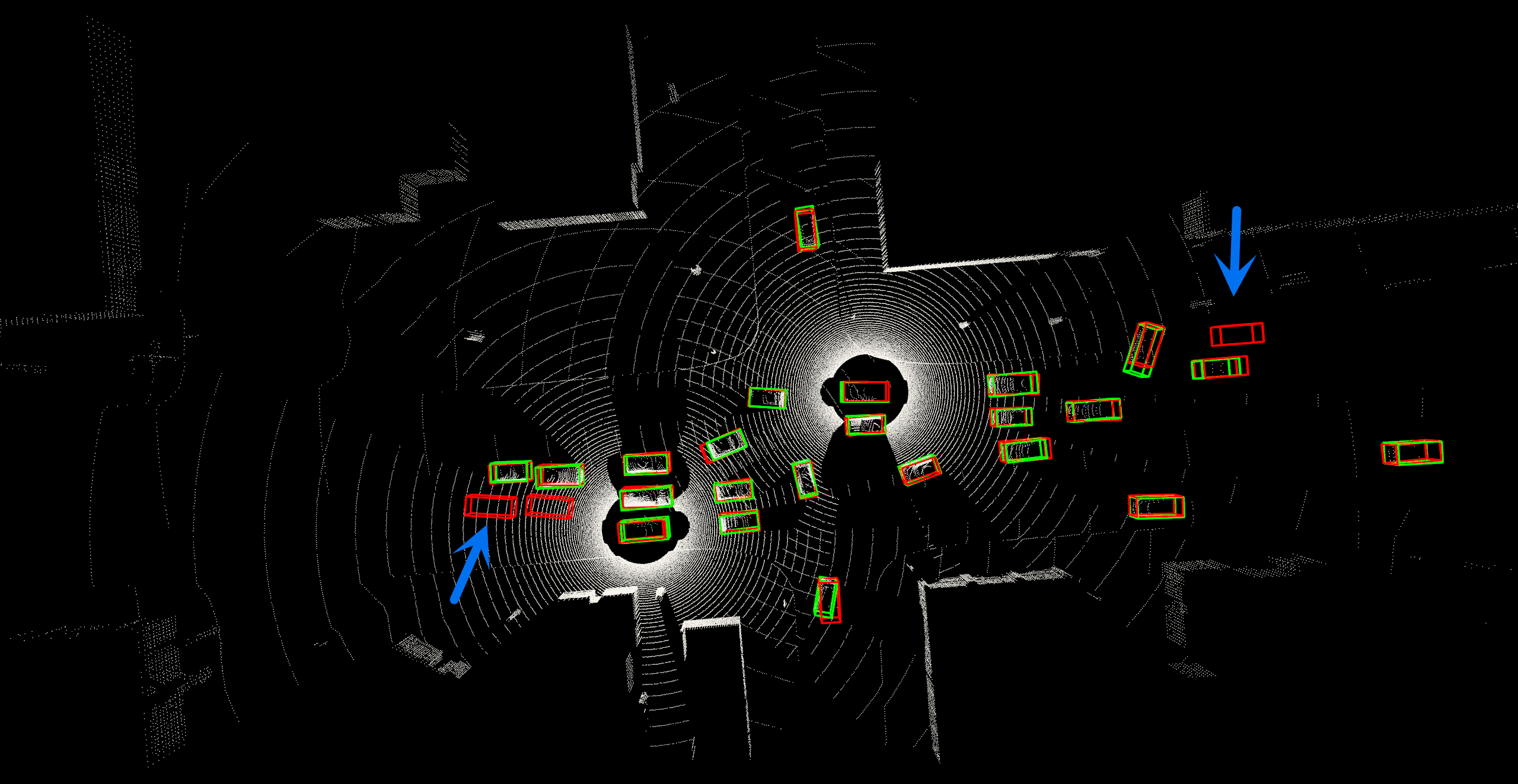}%
}
\subfigure[CoBEVT~\cite{xu2022cobevt}]{%
  \includegraphics[width=0.65\columnwidth]{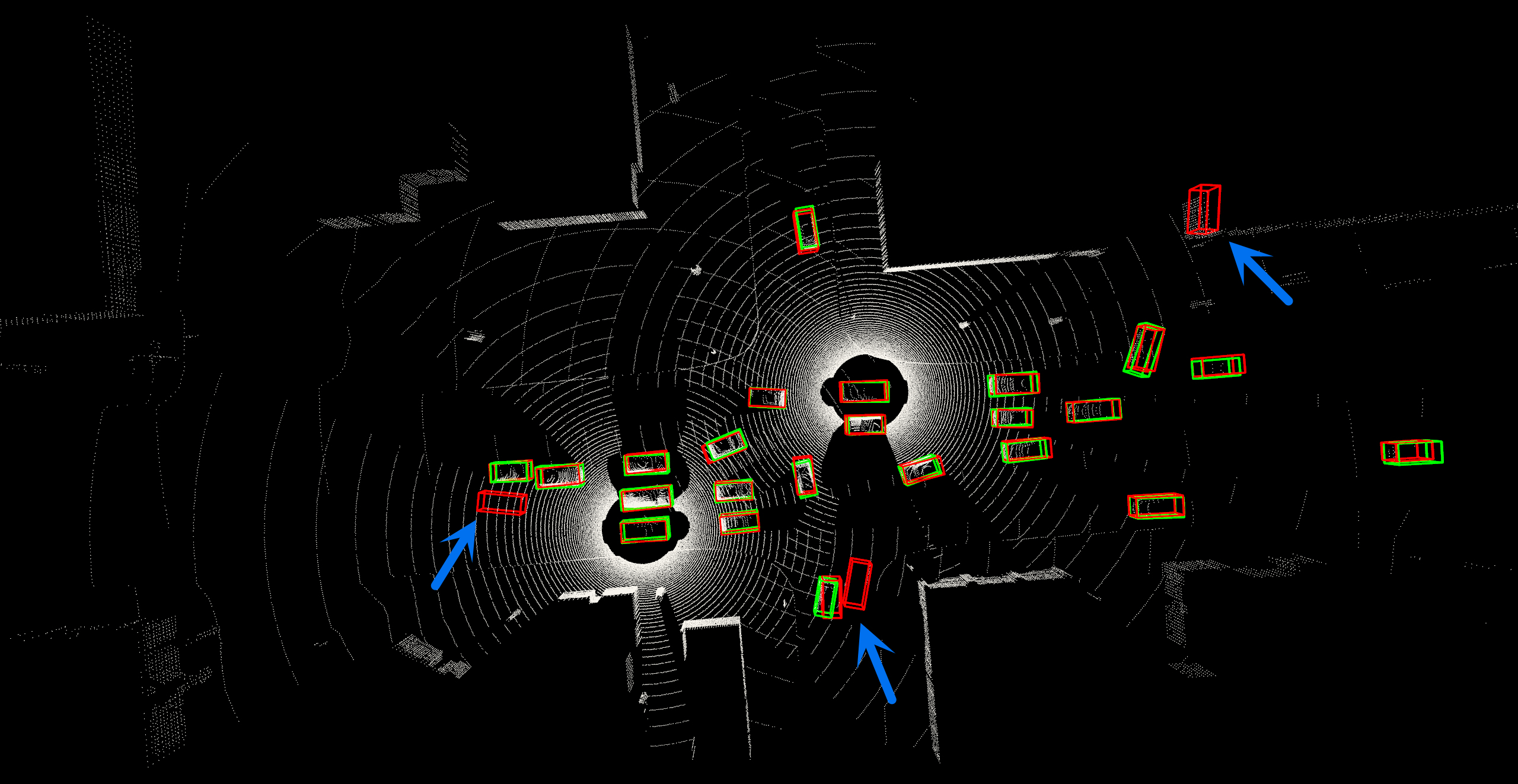}%
}
\subfigure[V2X-ViT~\cite{xu2022v2x}]{%
  \includegraphics[width=0.65\columnwidth]{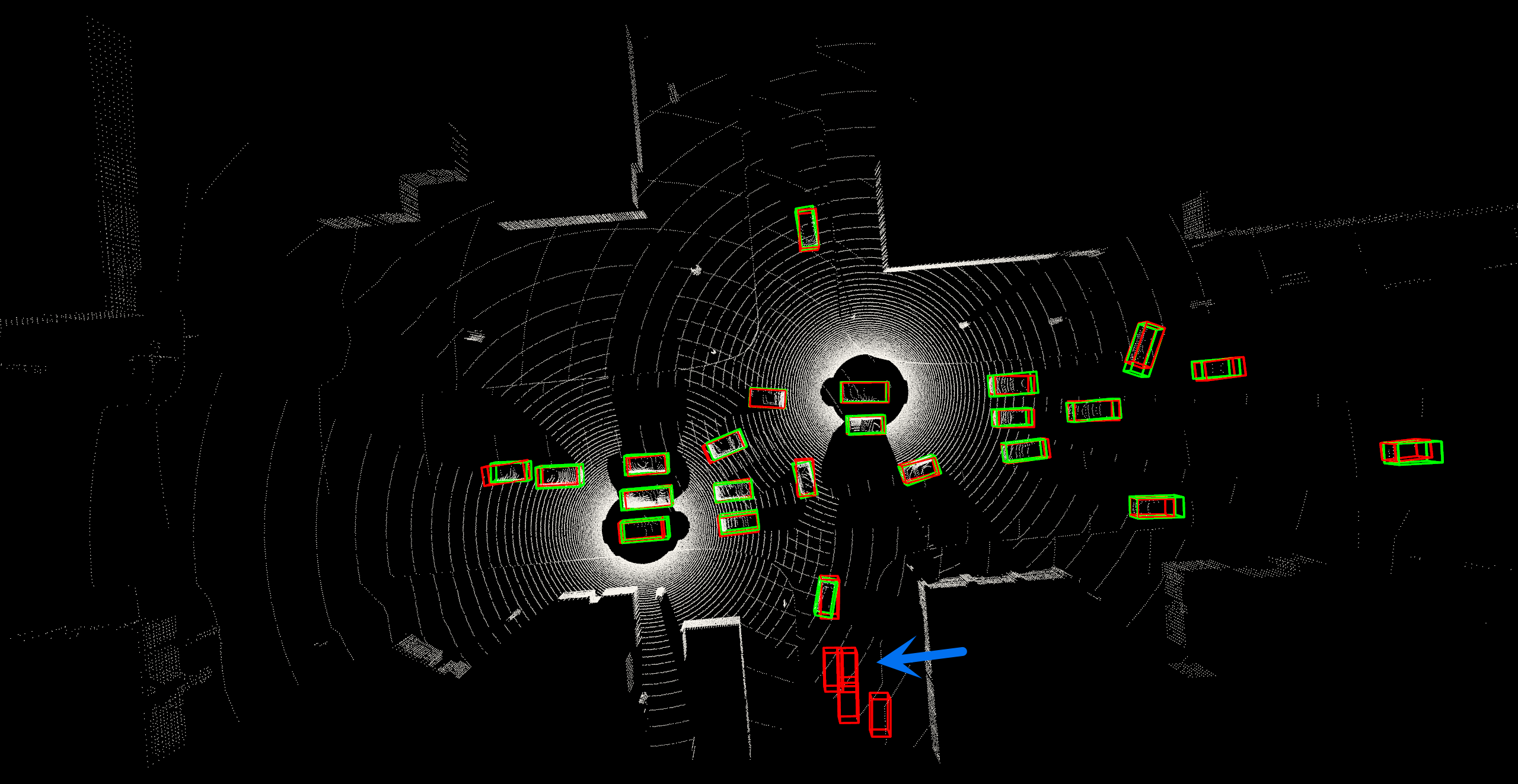}%
}
\subfigure[Proposed Method]{%
  \includegraphics[width=0.65\columnwidth]{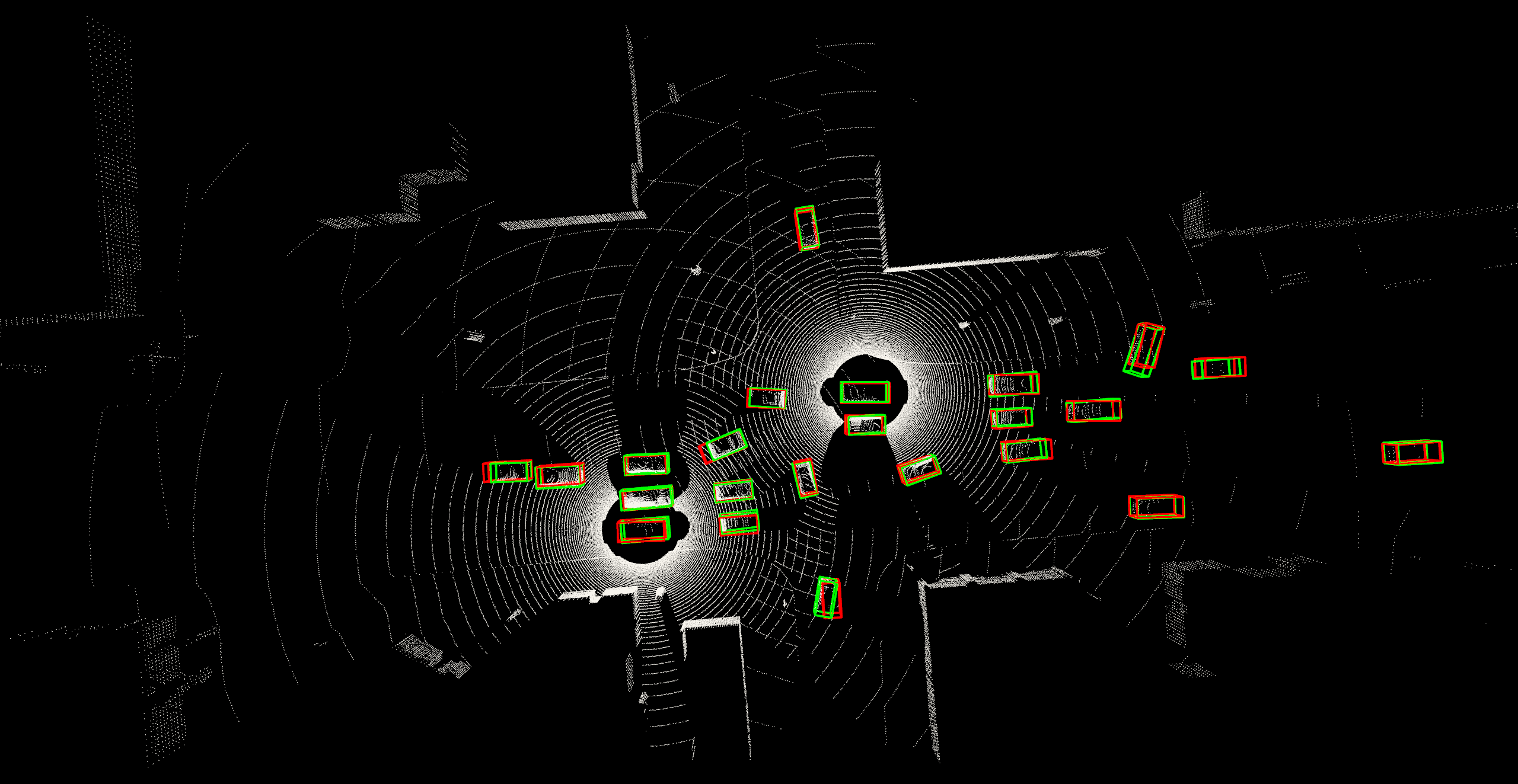}%
}
\caption{3D object detection visualization. \textcolor{green}{Green} and \textcolor{red}{red} 3D bounding boxes represent the \textcolor{green}{ground truth} and \textcolor{red}{prediction} respectively. The detection results of the proposed method are clearly more accurate. Some false detection examples are highlighted using blue arrow.}
\label{fig:visualization}
\end{figure*}

\begin{figure*}[!t] 
    \begin{centering}
        \includegraphics[width=1\textwidth]{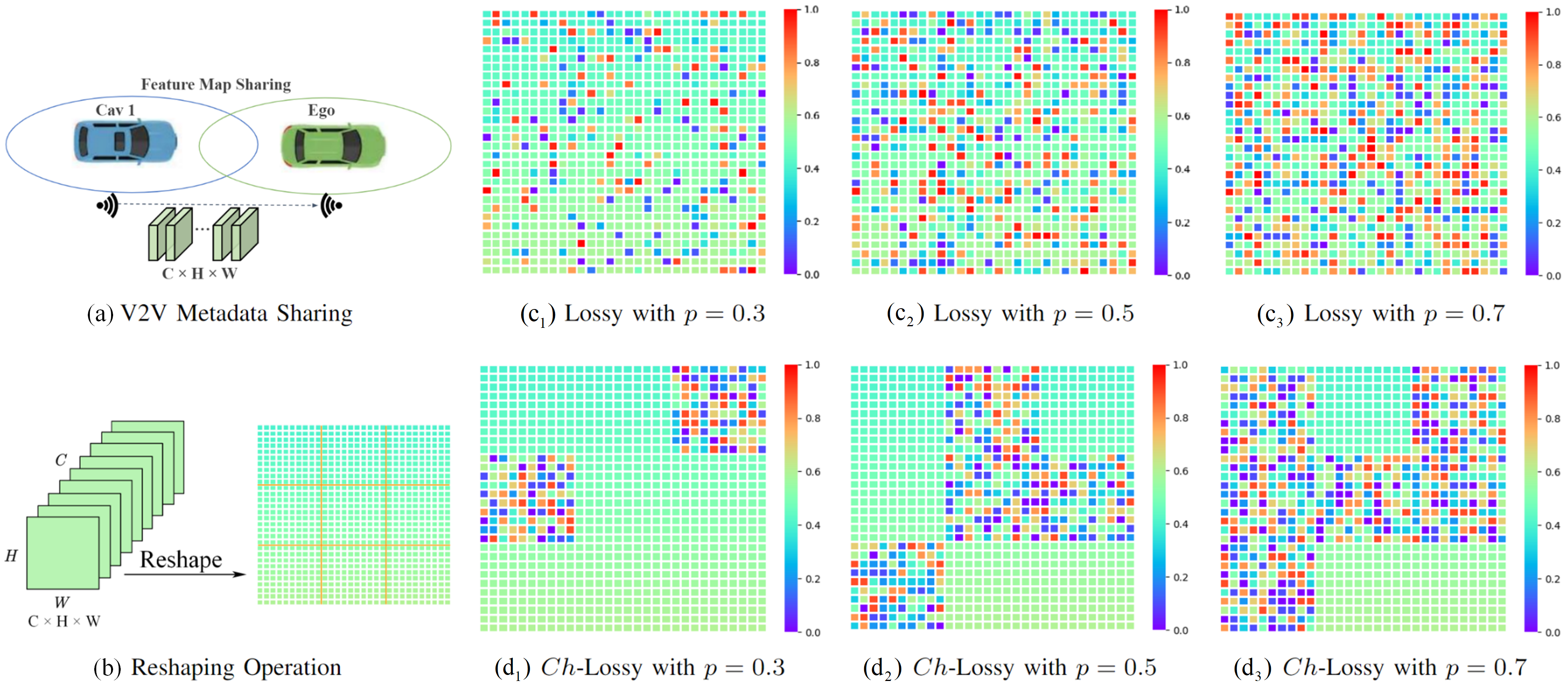}
    \par\end{centering}
    \caption{Illustration of different lossy communication types in V2V communication. (a) V2V Metadata Sharing, (b) Reshaping Operation, (c$_1$-c$_3$) Lossy Communication (named as ``Lossy") on the reshaped feature (b) with a global random selection probability $p$ of 0.3, 0.5, 0.7 respectively, (d$_1$-d$_3$) Channelwise Lossy Communication (named as ``$Ch$-Lossy") on the feature (b) with a channelwise random selection probability $p$ of 0.3, 0.5, 0.7 respectively. We use $C=9, H=10, W=10$ and normalized feature values for illustration in this example.}
    \label{fig:lossy}
\end{figure*}

\subsection{Experimental Results}

Table~\ref{tab:perfect} shows the performance comparisons of all models that are trained with \textit{Scheme I}, then tested on two communication types \textit{e.g.}, \textit{Ideal Communication} and \textit{Lossy Communication}, respectively. Under \textit{Ideal communication}, all the cooperative perception methods  significantly surpass \textit{NO Fusion} baseline. In V2V CARLA Town testing set, our proposed V2VAM outperforms the other five advanced fusion methods to achieve $92.6\%$/$86.1\%$ for AP@0.5/0.7, which is highlighted as bold text in Table~\ref{tab:perfect}. In V2V Culver City testing set, CoBEVT~\cite{xu2022cobevt} gets $87.7\%$/$74.8\%$ for AP@0.5/0.7, while the V2VAM achieves the $88.5\%$/$78.5\%$ for AP@0.5/0.7 as the best performance, which is higher than the second best fusion method CoBEVT~\cite{xu2022cobevt} with an AP@0.5/0.7 improvement of $1.6\%$/$3.7\%$. These results indicate cooperative perception methods can improve the perception performance than a single vehicle perception system under \textit{Ideal Communication}, and our proposed fusion method V2VAM can enhance the interaction between ego vehicle and other vehicles efficiently, which achieves the best performance.
However, under \textit{Lossy Communication} testing scenario, all intermediate fusion methods have a drastic performance drop on two testing sets, and the accuracy of these methods is even less than \textit{NO Fusion}. In V2V CARLA Town testing set, the cooperative perception performance of F-Cooper~\cite{chen2019f},  V2VNet~\cite{wang2020v2vnet}, OPV2V~\cite{xu2022opv2v}, and CoBEVT~\cite{xu2022cobevt} decrease by $80.8\%$, $85.0\%$, $85.6\%$, and $82.5\%$ in AP@0.5, respectively. Obviously, all intermediate fusion methods without considering the lossy communication are not practical for deployment in the real world.

The result of 3D object detection on two OPV2V testing sets based on the training of \textit{Scheme II} is presented in Table~\ref{tab:LC}. Under \textit{Lossy Communication}, although all intermediate fusion methods have a better performance than Table~\ref{tab:perfect}, which learned the lossy intermediate feature in the training stage. They still fail to handle lossy communication data resulting in the poor perception performance in ~\ref{tab:LC}. In V2V CARLA Town testing set, F-Cooper~\cite{chen2019f} got $49.2\%$,  V2VNet~\cite{wang2020v2vnet} got $46.5\%$, CoBEVT~\cite{xu2022cobevt} got $58.2\%$, and V2X-ViT~\cite{xu2022v2x} got $59.9\%$ in AP@0.7. These four fusion methods are even worse than single-vehicle baseline \textit{NO Fusion}, which indicates the highly negative impacts by lossy communication. While our proposed method
can reach the best performance of $84.1\%$/$70.5\%$ for AP@0.5/0.7 on V2V CARLA town testing set, and $84.6\%$/$66.3\%$ for AP@0.5/0.7 on Culver City testing sets, respectively. The proposed method achieves the best performance under both \textit{Ideal Communication} and  \textit{Lossy Communication}, which is highlighted in Table~\ref{tab:perfect}. Obviously, our proposed LCRN module efficiently maintains the benefits of collaborations under lossy communication. The proposed method can also diminish the impact of lossy V2V communication to achieve excellent cooperative perception performance. Further, we visualize some 3D object detection results on V2V Culver City testing set under \textit{Lossy Communication}, as shown in Fig.~\ref{fig:visualization}. Intuitively, these five comparison methods cannot handle loss communication appropriately, thus leading to some false negative proposals. While the proposed method improves the perception performance under lossy communication significantly.

\subsection{Discussion: Different Lossy Communication Types in V2V}
As explained in~\cite{nasralla2014subjective,belyaev2014robust}, several random issues such as the occurrence of obstacles, fast and changing vehicle speeds, distance between vehicles might result in lossy communication when sharing a set of communication data. To simulate the complex lossy communication in the real world, the sharing data is randomly selected by a uniformly distributed random probability $p \in [0,1]$  and then replaced by random noise within the range of original shared feature values. We design two ways of random selection to simulate different lossy communication types in the real-world V2V communication.

\textbf{Lossy Communication (named as ``Lossy") on global feature}: The shared feature after V2V metadata sharing is reshaped from 3D tensor to 2D matrix first (Fig.~\ref{fig:lossy}(b)). Then, as shown in Fig.~\ref{fig:lossy}(c$_1$-c$_3$), the reshaped feature is randomly selected by the global random probability $p$ and replaced by random noise within the range of original shared feature values.  

\textbf{Channelwise Lossy Communication (named as ``$Ch$-Lossy")}: Different with the ``Lossy" type to simulate lossy communication on the reshaped global feature, ``$Ch$-Lossy" type is to simulate lossy communication on different channels. As shown in as Fig.~\ref{fig:lossy}(d$_1$-d$_3$), given a shared feature $C \times H \times W$, $\lfloor p*C \rfloor$ channels are randomly selected by the channelwise random probability $p$ and replaced by random noise within the range of original shared feature values. 

Finally, the simulated lossy feature is reshaped back to its original shape of $C \times H \times W$ and then received by ego vehicle. In our experiment, \textbf{\textit{Scheme II}} utilizes the simulated lossy communication data by the ``Lossy" type to train models, and then we use the models trained in \textbf{\textit{Scheme II}} to test both ``Lossy" and ``$Ch$-Lossy" simulated data. Table~\ref{tab:two_types_lossy} shows the performance comparisons of several methods with the two lossy communication types. The proposed method achieves the best performance under both ``Lossy" and ``$Ch$-Lossy" communication types.

\begin{table}[htb]
\caption{3D detection performance comparison on two testing sets of OPV2V based on the training of \textit{Scheme II} with two different types of lossy communication.}
\label{tab:two_types_lossy}
\begin{tabular}{@{}cccccc@{}}
\toprule
\multirow{2}{*}{Method} &
  \multirow{2}{*}{\begin{tabular}[c]{@{}c@{}}Com.\\  Type\end{tabular}} &
  \multicolumn{2}{c}{V2V CARLA Towns} &
  \multicolumn{2}{c}{V2V Culver City} \\
 &
   &
  AP@ 0.5 &
  AP@ 0.7 &
  AP@ 0.5 &
  AP@ 0.7 \\ \midrule
OPV2V~\cite{xu2022opv2v} &
  \multicolumn{1}{c|}{\begin{tabular}[c]{@{}c@{}}Lossy\\ $Ch$-Lossy\end{tabular}} &
  \begin{tabular}[c]{@{}c@{}}0.739\\ 0.711\end{tabular} &
  \multicolumn{1}{c|}{\begin{tabular}[c]{@{}c@{}}0.603\\ 0.582\end{tabular}} &
  \begin{tabular}[c]{@{}c@{}}0.718\\ 0.737\end{tabular} &
  \begin{tabular}[c]{@{}c@{}}0.561\\ 0.582\end{tabular} \\ \midrule
CoBEVT~\cite{xu2022cobevt} &
  \multicolumn{1}{c|}{\begin{tabular}[c]{@{}c@{}}Lossy\\ $Ch$-Lossy\end{tabular}} &
  \begin{tabular}[c]{@{}c@{}}0.768\\ 0.742\end{tabular} &
  \multicolumn{1}{c|}{\begin{tabular}[c]{@{}c@{}}0.582\\ 0.615\end{tabular}} &
  \begin{tabular}[c]{@{}c@{}}0.795\\ 0.767\end{tabular} &
  \begin{tabular}[c]{@{}c@{}}0.586\\ 0.588\end{tabular} \\ \midrule
V2X-ViT~\cite{xu2022v2x} &
  \multicolumn{1}{c|}{\begin{tabular}[c]{@{}c@{}}Lossy\\ $Ch$-Lossy\end{tabular}} &
  \begin{tabular}[c]{@{}c@{}}0.770\\ 0.793\end{tabular} &
  \multicolumn{1}{c|}{\begin{tabular}[c]{@{}c@{}}0.599\\ 0.619\end{tabular}} &
  \begin{tabular}[c]{@{}c@{}}0.717\\ 0.731\end{tabular} &
  \begin{tabular}[c]{@{}c@{}}0.511\\ 0.520\end{tabular} \\ \midrule
Proposed &
  \multicolumn{1}{c|}{\begin{tabular}[c]{@{}c@{}}Lossy\\ $Ch$-Lossy\end{tabular}} &
  \textbf{\begin{tabular}[c]{@{}c@{}}0.841\\ 0.852\end{tabular}} &
  \multicolumn{1}{c|}{\textbf{\begin{tabular}[c]{@{}c@{}}0.705\\ 0.723\end{tabular}}} &
  \textbf{\begin{tabular}[c]{@{}c@{}}0.846\\ 0.851\end{tabular}} &
  \textbf{\begin{tabular}[c]{@{}c@{}}0.663\\ 0.675\end{tabular}} \\ \bottomrule
\end{tabular}
\end{table}

\subsection{Ablation Study}
The effectiveness of the two proposed components, V2VAM and LCRN, is investigated here. Based on training \textit{Scheme II}, all methods are evaluated under \textit{Lossy Communication} on  V2V CARLA Town and Culver City testing sets, respectively. AveFuse is used as the baseline fusion method, which just averages all intermediate features.  As shown in Table~\ref{tab:ablation_result}, the proposed V2VAM obtains $70.9\%$ in AP@0.5 and $58.3\%$ in AP@0.7 on V2V CARLA Town testing set, which is $7.7\%$ and $25.8\%$ higher than AveFusion in AP@0.5 and AP@0.7 respectively. 
Both Intra-vehicle attention and Inter-vehicle attention modules are quite effective for V2VAM if we remove one of them in V2VAM during the ablation study. By adding LCRN to the baseline method, AveFuse+LCRN achieves $69.8\%$ in AP@0.5 and $47.2\%$ in AP@0.7 on V2V CARLA Town testing set, with the improvement of $6.6\%$ in AP@0.5, and $14.7\%$ in AP@0.7. Furthermore, our proposed method V2VAM+LCRN achieves the best performance on both V2V CARLA Town and Culver City testing set. Obviously, both V2VAM and LCRN components are beneficial for improving the final performance of 3D object detection in lossy communication scenarios.

\section{Conclusions}\label{Sec:Conclusions}
In this paper, the side effect of lossy communication in the V2V cooperative perception is studied, and then we propose the first intermediate LC-aware feature fusion method considering lossy communication. An  LC-aware Repair Network (LCRN) is proposed to relieve the side effect of lossy communication and a specially designed V2V Attention Module (V2VAM) is designed to enhance the interaction between the ego vehicle and other vehicles including intra-vehicle attention of ego vehicle and uncertainty-aware inter-vehicle attention. The proposed method is verified in the digital-twin CARLA simulator based public cooperative perception dataset OPV2V, which is quite effective for the cooperative point cloud based 3D object detection under lossy V2V communication and outperforms other V2V point-cloud-based 3D object detection methods significantly.




 

%




\ifCLASSOPTIONcaptionsoff
  \newpage
\fi



%


\bibliographystyle{IEEEtran}
\bibliography{Jinlong}

\end{document}